\documentclass[11pt]{article}

\usepackage[preprint]{acl}

\usepackage{times}
\usepackage{latexsym}

\usepackage[T1]{fontenc}

\usepackage[utf8]{inputenc}

\usepackage{microtype}

\usepackage{inconsolata}

\usepackage{graphicx}

\usepackage{tabularx}
\usepackage{siunitx}
\usepackage{multirow}
\usepackage{booktabs}
\usepackage{microtype}
\usepackage{graphicx}
\usepackage{subfigure}
\usepackage{xcolor}         
\usepackage{tcolorbox}
\usepackage{makecell}
\usepackage{enumitem}
\newcommand{\tar}[1]{\texttt{{\color{red}{#1}}}}
\newcommand{\std}[1]{\small{$\pm$#1}}
\usepackage{hyperref}
\usepackage[hyphenbreaks]{breakurl}
\usepackage{flushend}
\usepackage{algorithm}
\usepackage{amsmath}
\usepackage{amssymb}
\usepackage{mathtools}
\usepackage{amsthm}
\usepackage[capitalize,noabbrev]{cleveref}

\newcommand{\benchmark}{UnFine}
\newcommand{\method}{FABLE}

\DeclareMathOperator*{\argmin}{arg\,min}

\title{FABLE: Fine-grained Fact Anchoring for Unstructured Model Editing}

\author{
 \textbf{Peng Wang\textsuperscript{1,2}},
 \textbf{Biyu Zhou\textsuperscript{1}\thanks{Corresponding authors.}},
 \textbf{Xuehai Tang\textsuperscript{1}},
\\
 \textbf{Jizhong Han\textsuperscript{1}},
 \textbf{Songlin Hu\textsuperscript{1,2}\footnotemark[1]},
\\
\\
 \textsuperscript{1}Institute of Information Engineering, Chinese Academy of Sciences \\
 \textsuperscript{2}School of Cyber Security, University of Chinese Academy of Sciences
\\
 \small{
   \textbf{Correspondence:} \href{mailto:email@domain}{\{wangpeng2022, zhoubiyu, tangxuehai, hanjizhong, husonglin\}@iie.ac.cn}
 }
}

\begin{document}
\maketitle
\begin{abstract}
Unstructured model editing aims to update models with real-world text, yet existing methods often memorize text holistically without reliable fine-grained fact access. To address this, we propose \textbf{\method}, a hierarchical framework that decouples fine-grained fact injection from holistic text generation. {\method} follows a two-stage, fact-first strategy: discrete facts are anchored in shallow layers, followed by minimal updates to deeper layers to produce coherent text. This decoupling resolves the mismatch between holistic recall and fine-grained fact access, reflecting the unidirectional Transformer flow in which surface-form generation amplifies rather than corrects underlying fact representations. We also introduce \textbf{\benchmark}, a diagnostic benchmark with fine-grained question–answer pairs and fact-level metrics for systematic evaluation. Experiments show that {\method} substantially improves fine-grained question answering while maintaining state-of-the-art holistic editing performance. Our code is publicly available at \url{https://github.com/caskcsg/FABLE}.
\end{abstract}

\section{Introduction}
Large language models (LLMs) have transformed natural language processing\cite{DBLP:conf/nips/BrownMRSKDNSSAA20, DBLP:conf/coling/HuangY000SZ22, liu2024deepseek}, yet static pre-training limits their capacity to absorb dynamically evolving factual knowledge\cite{ke, mend}. Fully retraining models to incorporate new information is computationally prohibitive and impractical in real-world settings\cite{DBLP:conf/emnlp/GuptaMS00WT23,DBLP:conf/emnlp/YaoWT0LDC023}. This has motivated growing interest in model editing, which aims to update specific knowledge by modifying a small subset of parameters while preserving overall model behavior\cite{DBLP:journals/csur/WangZLZCL25}. Among existing approaches, locate-then-edit methods such as ROME\cite{rome} and MEMIT\cite{memit} have shown notable success by identifying and modifying the causally relevant parameters to inject structured triple knowledge in the form of <subject, relation, object>.

Nevertheless, in real-world scenarios, approximately 80\% of knowledge is expressed as unstructured form \cite{DBLP:conf/wcre/Bavota16}, which introduces two key challenges for model editing: entity localization in open text and limited editing capacity for long content. Recent work on unstructured model editing, such as UnKE\cite{unke} and AnyEdit\cite{anyedit}, tackles these challenges by expanding the scope of editable parameters and anchoring optimization to context-preserving key tokens, thereby enabling holistic memorization and recall of unstructured text.

\begin{figure}[t]
    \centering
    \includegraphics[width=1.0\linewidth]{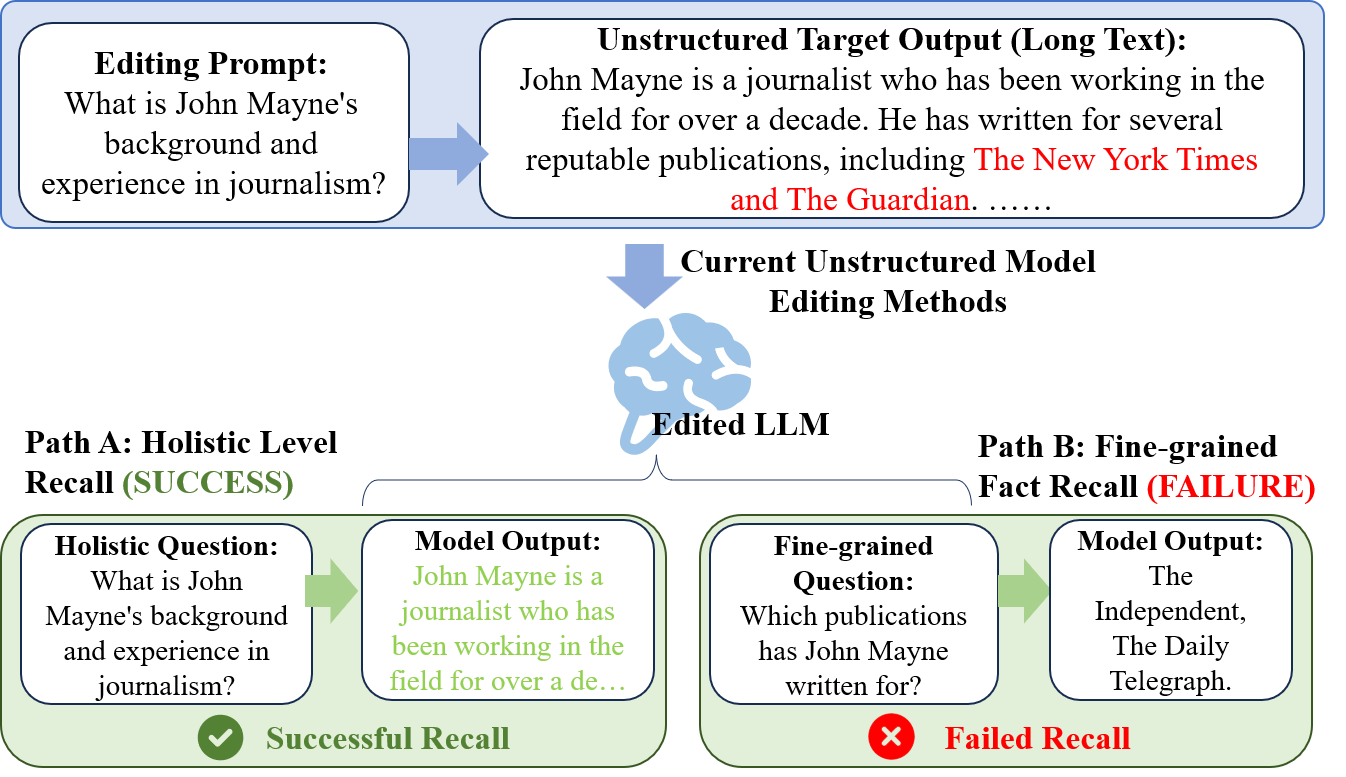}
    \caption{Limitation of existing unstructured editing methods (holistic recall vs. fine-grained fact access)}
    \label{fig:introduction}
\end{figure}


Despite these advances, we identify a fundamental limitation: existing methods enable \textbf{holistic  recall} of edited text, but fail to support \textbf{fine-grained fact access}, preventing reliable retrieval of atomic facts\cite{DBLP:conf/emnlp/MinKLLYKIZH23}. As shown in Figure~\ref{fig:introduction}, a model edited by UnKE can restate the entire text for the question "What is John Mayne's background and experience in journalism?", suggesting successful memorization at the textual level. However, when the query shifts to specific details within the text, the model fails to produce accurate answers. We attribute this limitation to the model primarily learning a high-level mapping from the question to surface-form representations for holistic recall (observable in layers associated with surface form generation), without consistently encoding the underlying atomic facts into lower-level knowledge storage. In real-world applications, downstream tasks often require fine-grained, targeted knowledge retrieval (e.g., querying specific facts in a report or particular imagery in a poem), rather than mere reproduction of edited text. 

Therefore, we propose \textbf{\method}, a novel framework for unstructured model editing. Unlike existing methods that primarily optimize surface-form recall at the holistic text level, {\method} adopts a two-stage hierarchical strategy that decouples fine-grained facts from holistic textual surface forms and anchors them into distinct parameter layers. 
Motivated by the "early decoding" phenomenon in Transformers, {\method} adopts a fact-first, generation-later design, embedding fine-grained facts into shallow layers before surface-form realization. Editing proceeds in two stages: discrete facts are first injected into shallow fine-grained key generators via synthesized question–answer pairs, from our \textbf{\benchmark} benchmark, then minimal, localized adjustments are applied to deeper surface-form key generators to ensure global coherence and fluent generation. This hierarchically decoupled design enables the model to achieve both reliable holistic surface-form recall and accurate fine-grained factual access to the edited content.

In summary, the  main contributions are:

(1) We identify a mismatch in unstructured editing between holistic surface-form memorization and fine-grained fact recall.

(2) We introduce {\benchmark}, a diagnostic benchmark for fine-grained fact recall, together with tailored metrics (e.g., $HR$, $C_{\text{LCS}}$). 

(3) We propose {\method}, a hierarchical editing framework that decouples fine-grained fact anchoring from surface-form generation.

\section{Benchmark Construction}
\label{section:benchmark_construct}
Existing benchmarks evaluate if a model outputs $A$ given $Q$, but not whether it captures the fine-grained fact within $A$. To close this gap, we introduce {\benchmark}, which extends existing benchmarks by incorporating fine-grained factual question answering as a core evaluation dimension.

\subsection{Datasets}
{\benchmark} is built upon three widely-used datasets in the field of unstructured model editing: UnKEBench\cite{unke}, AKEW (CounterFact)\cite{akew}, and AKEW (MQuAKE)\cite{akew}. Among them, UnKEBench is the first benchmark specifically designed for evaluating unstructured model editing capabilities in LLMs, comprising complex unstructured long-text question-answer pairs. AKEW is the first large-scale unstructured model editing benchmark constructed from real-world scenarios; its subsets based on CounterFact\cite{rome} and MQuAKE\cite{mquake} are referred to as AKEW (CounterFact) and AKEW (MQuAKE), respectively.

We have made improvements in the following two aspects: \textbf{(1) Generation of Fine-Grained QA Pairs}: 
Noting that while UnKEBench includes natural language-based fine-grained QA pairs, AKEW (CounterFact) and AKEW (MQuAKE) lack such data. Following the methodology of UnKEBench, we augment the latter two with corresponding fine-grained QA pairs. \textbf{(2) Key Knowledge Phrase Extraction}: However, due to the nature of natural language, answers in these QA pairs contain not only factual knowledge but also certain linguistic styles. To more precisely evaluate whether the model has acquired the key knowledge (mitigating interference from language style), we further extract key knowledge phrases from each fine-grained answer. The improved datasets are renamed \textbf{\benchmark-UnKE}, \textbf{\benchmark-CF}, and \textbf{\benchmark-MQ}, respectively. Detailed dataset construction procedures are provided in Appendix~\ref{appendix:datasets_construction}.

\subsection{Evaluation Metrics}
We use a two-level evaluation framework to assess both holistic and fine-grained consistency between model outputs and target edits.

\textbf{(1) Holistic Knowledge QA Evaluation.}
Following prior works\cite{unke,anyedit}, we adopt lexical similarity (ROUGE-L\cite{rouge}) and semantic similarity (BERT-Score\cite{bertscore}) to evaluate the editing success rate.

\textbf{(2) Fine-Grained Knowledge QA Evaluation.}
Beyond sentence-level similarity, we introduce the following two fact-level metrics to assess the model’s ability to recall fine-grained fact.

\textbf{Hit Rate (HR):} It evaluates precise fact recall by checking, via exact string matching, whether the model output contains all key knowledge phrases extracted from the gold-standard answer. 

\textbf{Longest Common Subsequence Coverage ($C_{\text{LCS}}$):} This metric quantifies the completeness of content coverage. It is calculated as the ratio of the length of the Longest Common Subsequence (LCS) between the model output and the fine-grained gold answer to the total word count of the fine-grained gold answer. It effectively measures the output's ability to capture key information points. For example, for a answer "\textbf{North} Island \textbf{of} New \textbf{Zealand}" and a model output "\textbf{North of Zealand}", the LCS is "North", "of", "Zealand" (length 3), yielding a coverage of $C_{\text{LCS}} = \frac{3}{5} = 0.6$. Formal definitions and detailed formulations for all evaluation metrics are provided in Appendix~\ref{appendix:evaluation_metrics}.

\begin{figure*}[t]
    \centering
    \includegraphics[width=0.75\linewidth]{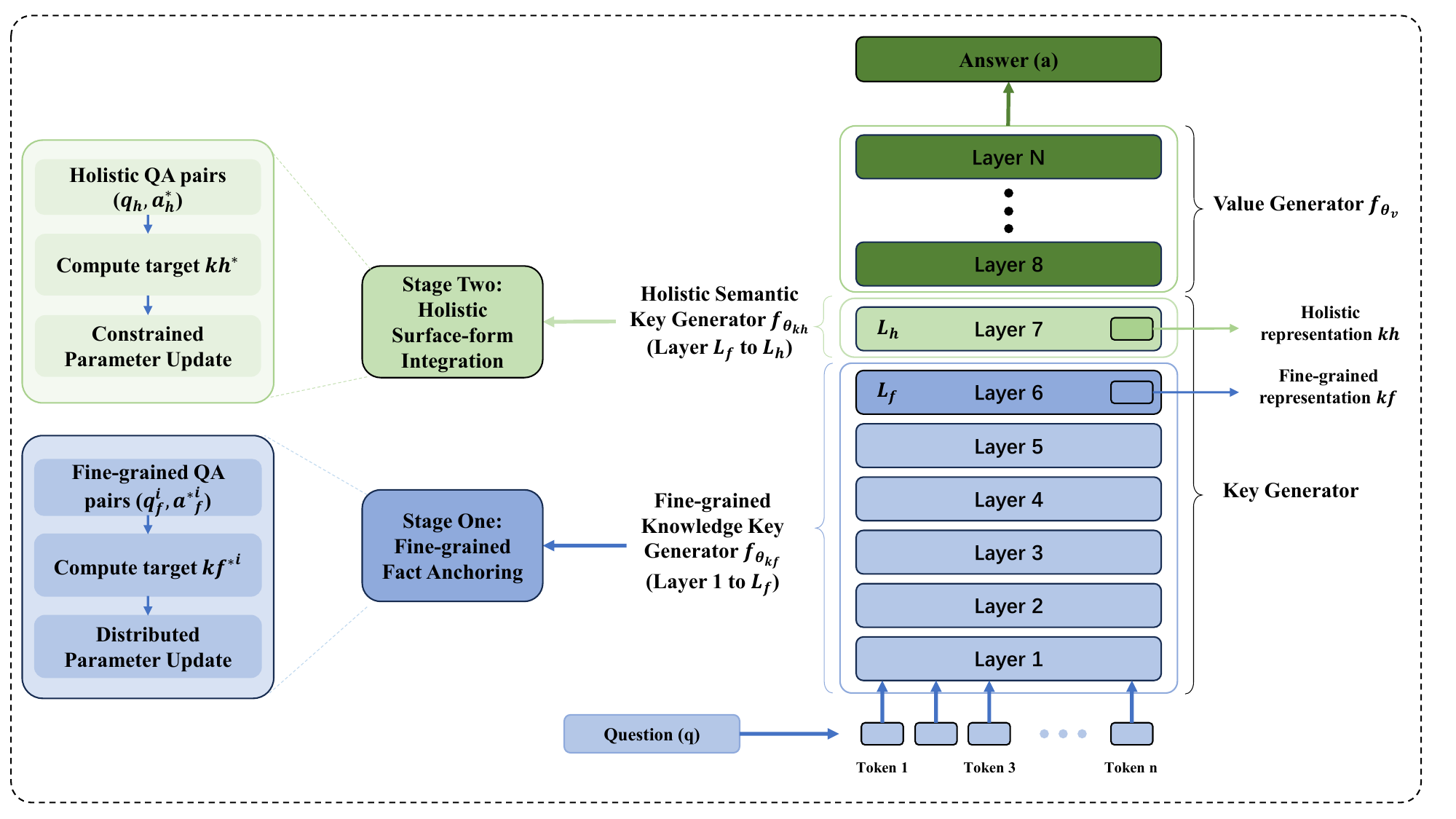}
    \caption{FABLE decomposes the key generator in a Transformer-based LLM into a two-stage hierarchical process: (1) fine-grained fact anchoring, which first encodes and stabilizes discrete factual knowledge; and (2) holistic surface-form integration, which then organizes these facts into a coherent narrative. The architecture consists of a fine-grained key generator, a holistic key generator, and a value generator.}
    \label{fig:method}
\end{figure*}

\section{FABLE}
\subsection{Hierarchical Key-Value Storage Architecture}
\label{section:arch}
Existing research often views Transformer-based LLMs as continuous key-value memory networks~\cite{unke}. In this view, shallow layers are seen as a key generator, responsible for compressing input information to produce knowledge representations (keys); while the middle to deep layers act as a value generator, responsible for decoding the keys and injecting them into the residual stream to form value vectors. This block-level key-value partitioning retains the attention mechanism and non-linear transformations, making it more suitable for representing unstructured knowledge compared to traditional MLP-level key-value pairs\cite{rome,memit}.

However, the prevailing view treats the key generator as a monolithic operation, which can lead to overfitting surface forms while failing to reliably encode fine-grained facts. For unstructured text, generating an effective "key" should instead be decomposed into two levels: (1) \textbf{fine-grained fact anchoring}, which extracts and stabilizes discrete factual units, and (2) \textbf{holistic surface-form construction}, which organizes these units into a coherent, fluent narrative. This decomposition leverages the natural stratification of representations in Transformers, where lower layers excel at capturing local, fine-grained features and higher layers integrate them into global semantic representations.

To formalize this, we first represent an N-layer 
model as $f_{\theta}=f^1_{\theta_1} \circ f^2_{\theta_2} \circ \cdots \circ f^N_{\theta_N}$, where $\theta$ denotes the model parameters, $\circ$ denotes layer composition, and $f^l_{\theta_l}$ represents the $l$-th layer and its parameters $\theta_l$. We formalize $f_{\theta}$ as a combination of two core modules: an Unstructured Knowledge Key Generator($G_{K}$) and a Value Generator($G_{V}$). The key generator can be further decomposed into two consecutive stages:
\begin{equation}
f_{\theta} = \underbrace{\left( \mathcal{F}_{\text{fine}} \circ \mathcal{F}_{\text{hol}} \right)}_{\text{$G_{K}$}} \circ \underbrace{\mathcal{V}.}_{\text{$G_{V}$}}
\end{equation}

Here, the fine-grained key generator $\mathcal{F}_{\text{fine}}$ encodes layers $1$ to $L_f$. The holistic key generator 
$\mathcal{F}_{\text{hol}}$ encodes layers $L_{f+1}$ to $L_h$. 
Finally, the value generator $\mathcal{V}$ maps layers $L_{h+1}$ to $N$ into the token space. Here, $L_f$ and $L_h$ denote the split layer between fine-grained and holistic key generators, and the boundary separating key and value generators, respectively.

Given a question $q=[q_{1},\dots,q_{n}]$ (where $n$ is the token count of $q$), its representation at layer $l$ is $h^l_{q} = [h^l_{q,1},\dots,h^l_{q,n}]$, and its forward propagation process is:
\begin{equation}
h^l_{q} = f^l_{\theta_l}(h^{l-1}_{q}),
\end{equation}
where $h^1_{q} = f^1_{\theta_1}(h^{0}_{q}) = f^1_{\theta_1}(q)$. For convenience in subsequent discussion, we denote the representation corresponding to the $i$-th token at layer $l$ as: $h^l_{q,i} = h^l_{q}[i] = f^l_{\theta_l}(h^{l-1}_{q})[i]$. Since the vector corresponding to the last token typically highly aggregates information from the entire $q$, we further define the fine-grained fact key $k_\text{fine}$, the holistic semantic key $k_\text{hol}$, and the value $v$ as:
\begin{gather}
k_\text{fine} = \mathcal{F}_{\text{fine}}(q)[n].\\
k_\text{hol} = \mathcal{F}_{\text{hol}}([h^{L_f}_{q,1},\dots,h^{L_f}_{q,n-1},k_\text{fine}])[n].\\
v = \mathcal{V}([h^{L_h}_{q,1},\dots,h^{L_h}_{q,n-1},k_\text{hol}])[n].
\end{gather}

Finally, the model decodes based on $v$ to obtain the output answer $a=[a_{1},\dots,a_{m}]$ (where $m$ is the token count of $a$).

\subsection{Stage One: Fine-grained Fact Anchoring}
\label{section:fine_inject}
This stage aims to inject fine-grained facts into the model parameters, i.e., to update the parameters of $\mathcal{F}_{\text{fine}}$ so that the model can accurately output the target answer $a^*_f$ when presented with a fine-grained question $q_f$. The specific implementation is divided into two steps: first, compute the target fine-grained fact key $k_\text{fine}^*$ that can guide the model to generate $a^*_f$; then, adjust the parameters of $\mathcal{F}_{\text{fine}}$ so that it stably produces $k_\text{fine}^*$ when input with $q_f$.

\paragraph{(1) Computing the Target Fine-Grained Fact Key}
We search for the optimal direction in the residual stream at layer $L_f$ that can trigger the target fact. By optimizing the residual vector $\delta_f$ to minimize the negative log probability of the target answer $a^*_f$, we obtain  $\delta_f$:  
\begin{gather}
\label{equation:delta_f}
\delta_f = \argmin_{\delta_f}-\log P_{f_\theta(k_\text{fine}\,\mapsto\,k_\text{fine}+\delta_f)}(a_f^*\mid q_f)\\
k_\text{fine}^* = k_\text{fine} + \delta_f,
\end{gather}
where $f_\theta(k_\text{fine}\mapsto k_\text{fine}+\delta_f)$ denotes replacing $k_\text{fine}$ with $k_\text{fine}+\delta_f$ during forward propagation. 

\paragraph{(2) Distributed Parameter Update}
Since $\mathcal{F}_{\text{fine}}$ involves multi-layer continuous transformations, to ensure consistency of the bottom-up residual flow, we distribute the parameter update across its layers ($l \in [1, L_f]$). We define a per-layer optimization objective for each layer, making it share a portion of the total offset $\delta_f$. Specifically, for layer $l$, the objective is to make its output at the last token position ${h^*}^l_{q_f,n} = f^l_{\theta^*_l}(h^{l-1}_{q_f})[n]$ approximate $h^l_{q_f,n} + \frac{\delta_f}{L_f - l + 1}$, where $\theta^*_l$ represents the updated parameters for layer $l$. In particular, when $l = L_f$, we have: $h^{L_f}_{q_f,n} + \frac{\delta_f}{L_f - L_f + 1} = k_\text{fine} + \delta_f = k_\text{fine}^*$.

To preserve the model's behavior on the input prefix, the representations of the first $n-1$ tokens are kept unchanged. The layer-$l$ optimization objective is:

\begin{equation}
\scalebox{0.9}{$\displaystyle
\begin{aligned}
    \theta^*_l = \argmin_{\theta^*_l} \Big( & {\left\| {h^*}^l_{q_f,n} - h^l_{q_f,n} - \frac{\delta_f}{L_f - l + 1} \right\|^2} \\
    & + {\sum_{j=1}^{n-1} \left\| {h^*}^l_{q_f,j} - h^l_{q_f,j} \right\|^2} \Big).
\end{aligned}
$}
\end{equation}

Since unstructured text often contain more than a piece of fine-grained fact, we consider $u$ different fine-grained questions $\{q_f^i\}_{i=1}^u$, each $q_f^i$ contains $n$ tokens (for the method of extracting multi-aspect fine-grained QA pairs $(q^i_f,{a^*}^i_f)$ from unstructured text, see Appendix~\ref{appendix:multi-aspect-extraction}). Additionally, to preserve the model's predictions on irrelevant samples as much as possible while editing the target knowledge, we introduce v irrelevant samples $D=\{d^i\}_{i=1}^v$, each $d^i$ contains $w$ tokens. Therefore, the complete optimization objective for layer $l$ is:
\begin{equation}
\scalebox{0.9}{$\displaystyle
\begin{aligned}
\label{equation:fine_final}
    \theta^*_l = \argmin_{\theta^*_l} \Big( & \underbrace{\sum_{i=1}^{u} \left\| {h^*}^l_{q^i_f,n} - h^l_{q^i_f,n} - \frac{\delta^i_f}{L_f-l+1} \right\|^2}_{\text{Edit Efficacy}} \\
    & + \underbrace{\sum_{i=1}^{u}\sum_{j=1}^{n-1} \left\| {h^*}^l_{q^i_f,j} - h^l_{q^i_f,j} \right\|^2}_{\text{Prefix Consistency}} \\
    & + \underbrace{\sum_{k=1}^{v}\sum_{j=1}^{w} \left\| {h^*}^l_{d^k,j} - h^l_{d^k,j} \right\|^2}_{\text{Locality Preservation}} \Big).
\end{aligned}
$}
\end{equation}

By optimizing Eq.~\ref{equation:fine_final} layer-by-layer, we obtain the updated $\mathcal{F}_{\text{fine}}^*$ with $\mathcal{F}_{\text{fine}}^* = f^1_{\theta_1^*}\circ f^2_{\theta_2^*}\circ\cdots\circ f^{L_f}_{\theta_{L_f}^*}$.

\subsection{Stage Two: Holistic Surface-form Integration}
\label{section:holistic}
After embedding fine-grained fact into $\mathcal{F}_{\text{fine}}^*$, we further adjust the parameters of $\mathcal{F}_{\text{hol}}$ to perform semantic integration for a holistic question $q_h$, in order to generate fluent unstructured narrative $a^*_h$. According to prior work\cite{unke}, updating only the single layer $L_h$ parameters can effectively enable the model to narrate unstructured knowledge, so we maintain this setting. Similarly, the implementation of this stage is also divided into two steps. First, analogous to Eq.~\ref{equation:delta_f}, we can obtain the residual vector $\delta_h$ and the target semantic key $k_\text{hol}^*$ via optimization. Then, analogous to Eq.~\ref{equation:fine_final}, we can formulate the optimization objective for $\mathcal{F}_{\text{hol}}^*$, but with three key differences: (1) Only update the single layer $L_h$; (2) The QA pair is a single $(q_h, a^*_h)$ rather than a batch; (3) To prevent blindly updating the semantic layer and overwriting the fine-grained fact signals passed from the first stage, a fine-grained preservation constraint is introduced. Therefore, the optimization objective is:
\begin{equation}
\scalebox{0.9}{$\displaystyle
\begin{aligned}
\label{equation:holistic_final}
    \theta^*_{L_h} = \argmin_{\theta^*_{L_h}} \Big( & \underbrace{ \left\| {h^*}^{L_h}_{q_h,n} - h^{L_h}_{q_h,n} - k_\text{hol}^* \right\|^2}_{\text{Edit Efficacy}} \\
    & + \underbrace{\sum_{j=1}^{n-1} \left\| {h^*}^{L_h}_{q_h,j} - h^{L_h}_{q_h,j} \right\|^2}_{\text{Prefix Consistency}} \\
    & + \underbrace{\sum_{k=1}^{v}\sum_{j=1}^{w} \left\| {h^*}^{L_h}_{d^k,j} - h^{L_h}_{d^k,j} \right\|^2}_{\text{Locality Preservation}} \\
    & + \underbrace{\sum_{i=1}^{u}\sum_{j=1}^{n} \left\| {h^*}^{L_h}_{q^i_f,j} - h^{L_h}_{q^i_f,j}  \right\|^2}_{\text{Fine-grained Preservation}} \Big).
\end{aligned}
$}
\end{equation}

\subsection{Implementation Details}
\label{section:imple}
In the concrete implementation, to balance editing effectiveness and model stability, we designed the parameter update scope and data scale specifically. First, for updating the fine-grained fact injection stage $\mathcal{F}_{\text{fine}}$, we do not update all layers 1 to $L_f$. Because the initial layers of the model primarily handle basic functions like syntactic understanding. Therefore, we choose to update only the middle layers 4, 5, and 6. Second, for updating the holistic semantic integration stage $\mathcal{F}_{\text{hol}}$, following prior work, we set $L_h$ as 7 and update only its parameters. Regarding data, the number of QA pairs used for fine-grained fact injection is set to 5 times the number of seed QA pairs $S$ extracted from the unstructured text in the editing sample, i.e., $5 \times S$ (details on seed QA pairs are in Appendix~\ref{appendix:multi-aspect-extraction}). Meanwhile, to preserve the model's general capabilities as much as possible during editing, following related research settings, for each editing sample, we randomly sample 20 samples from the Alpaca instruction-tuning dataset\footnote{\href{https://github.com/tatsu-lab/stanford_alpaca}{https://github.com/tatsu-lab/stanford\_alpaca}} to serve as the irrelevant sample set $D$. For ablation studies on the choice of layers and number of QA pairs, see Section~\ref{section:ablation_study}.

\section{Experiments}

\begin{table*}[t]
\centering
\caption{Unstructured model editing performance with different models, methods, and datasets. “Pre-edited” denotes the original model before editing; “–” indicates that the corresponding field is absent from the dataset. Best results are \textbf{boldfaced}; second-best are \underline{underlined}.}
\renewcommand{\arraystretch}{0.6} 
\resizebox{0.85\textwidth}{!}{%
\begin{tabular}{c|c|cc|cccc}
\toprule[1.5pt]

\multirow{2}{*}{\textbf{Model}} & \multirow{2}{*}{\textbf{Method}} & \multicolumn{2}{c|}{\textbf{Holistic}} & \multicolumn{4}{c}{\textbf{Fine-grained}} \\ 
\cmidrule(lr){3-8}

& & \textbf{Bert-Score$\uparrow$} & \multicolumn{1}{c|}{\textbf{Rouge-L$\uparrow$}} &\textbf{Bert-Score$\uparrow$} & \textbf{Rouge-L$\uparrow$} & \textbf{HR$\uparrow$} & \textbf{\(C_{\text{LCS}}\)$\uparrow$}\\ 
\midrule[1.0pt]

\multicolumn{8}{c}{\textbf{\benchmark-UnKE}} \\
\midrule[1.0pt]

\multirow{7}{*}{\rotatebox{90}{{\textbf{Llama3-8B}}}} 
&\textbf{Pre-edited}& {71.83\std{0.11}} & {22.69\std{0.10}} & {22.26\std{0.08}} & {12.81\std{0.09}} & {5.00\std{0.10}} & {17.50\std{0.12}}\\
&\textbf{FT-L}& {33.29\std{0.22}} & {13.35\std{0.12}} & {26.10\std{0.10}} & {15.49\std{0.11}} & {7.60\std{0.12}} & {21.83\std{0.14}}\\
&\textbf{ROME}& {79.51\std{0.13}} & {40.00\std{0.23}} & {24.76\std{0.09}} & {17.79\std{0.12}} & {9.53\std{0.14}} & {24.70\std{0.15}}\\
&\textbf{MEMIT}& {81.78\std{0.20}} & {49.64\std{0.29}} & {25.60\std{0.09}} & {20.54\std{0.13}} & {12.51\std{0.16}} & {28.30\std{0.17}}\\
&\textbf{UnKE}& \underline{98.20\std{0.06}} & \underline{93.39\std{0.16}} & {27.93\std{0.09}} & {27.15\std{0.16}} & {20.55\std{0.20}} & {37.13\std{0.19}}\\
&\textbf{AnyEdit}& {96.06\std{0.07}} & {90.34\std{0.15}} & \underline{28.61\std{0.09}} & \underline{29.22\std{0.15}} & \underline{23.69\std{0.20}} & \underline{40.51\std{0.18}}\\
 &\textbf{\method}& \textbf{99.36\std{0.04}} & \textbf{97.78\std{0.10}} & \textbf{65.63\std{0.16}} & \textbf{53.79\std{0.19}} & \textbf{53.31\std{0.23}} & \textbf{61.78\std{0.19}}\\

\midrule[1pt]
\midrule[1pt]

\multirow{7}{*}{\rotatebox{90}{{\textbf{Qwen2.5-7B}}}}
&\textbf{Pre-edited}& {72.96\std{0.10}} & {23.74\std{0.10}} & {23.75\std{0.09}} & {13.21\std{0.10}} & {3.93\std{0.09}} & {17.74\std{0.11}}\\
&\textbf{FT-L}& {28.11\std{0.24}} & {14.19\std{0.10}} & {23.34\std{0.10}} & {14.23\std{0.10}} & {4.44\std{0.10}} & {21.39\std{0.12}}\\
&\textbf{ROME}& {78.53\std{0.13}} & {39.19\std{0.18}} & {25.59\std{0.09}} & {18.21\std{0.12}} & {8.52\std{0.13}} & {24.26\std{0.14}}\\
&\textbf{MEMIT}& {81.44\std{0.16}} & {49.92\std{0.24}} & {25.33\std{0.09}} & {19.65\std{0.13}} & {11.13\std{0.15}} & {26.53\std{0.16}}\\
&\textbf{UnKE}& {96.86\std{0.07}} & {90.24\std{0.17}} & {26.89\std{0.09}} & {22.47\std{0.13}} & {12.86\std{0.16}} & {30.10\std{0.16}}\\
&\textbf{AnyEdit}& \underline{97.02\std{0.07}} & \underline{91.95\std{0.17}} & \underline{27.75\std{0.09}} & \underline{25.47\std{0.15}} & \underline{17.63\std{0.17}} & \underline{34.74\std{0.17}}\\
 &\textbf{\method}& \textbf{98.86\std{0.04}} & \textbf{97.35\std{0.09}} & \textbf{50.87\std{0.16}} & \textbf{38.31\std{0.18}} & \textbf{31.35\std{0.21}} & \textbf{44.58\std{0.18}}\\
 
\midrule[1.0pt]

\multicolumn{8}{c}{\textbf{\benchmark-CF}} \\
\midrule[1.0pt]

\multirow{7}{*}{\rotatebox{90}{{\textbf{Llama3-8B}}}} 
&\textbf{Pre-edited}& {71.42\std{0.12}} & {15.53\std{0.09}} & {26.37\std{0.09}} & {26.56\std{0.17}} & {20.22\std{0.19}} & {32.43\std{0.19}}\\
&\textbf{FT-L}& {29.71\std{0.20}} & {10.19\std{0.09}} & {30.12\std{0.10}} & {25.80\std{0.17}} & {20.27\std{0.19}} & {31.86\std{0.19}}\\
&\textbf{ROME}& {80.38\std{0.15}} & {40.33\std{0.23}} & {27.89\std{0.08}} & {30.40\std{0.17}} & {24.89\std{0.21}} & {38.01\std{0.19}}\\
&\textbf{MEMIT}& {83.74\std{0.18}} & {54.10\std{0.30}} & {28.64\std{0.08}} & {32.88\std{0.18}} & {29.25\std{0.21}} & {41.48\std{0.20}}\\
&\textbf{UnKE}& \textbf{99.28\std{0.04}} & \underline{97.02\std{0.12}} & {29.22\std{0.08}} & {35.89\std{0.19}} & {32.64\std{0.22}} & {45.62\std{0.20}}\\
&\textbf{AnyEdit}& \underline{98.19\std{0.05}} & {94.97\std{0.11}} & \underline{30.41\std{0.08}} & \underline{39.05\std{0.19}} & \underline{38.94\std{0.22}} & \underline{51.40\std{0.19}}\\
&\textbf{\method}& \textbf{99.28\std{0.05}} & \textbf{98.35\std{0.09}} & \textbf{71.89\std{0.13}} & \textbf{64.59\std{0.19}} & \textbf{66.62\std{0.22}} & \textbf{73.27\std{0.18}}\\

\midrule[1pt]
\midrule[1pt]

\multirow{7}{*}{\rotatebox{90}{{\textbf{Qwen2.5-7B}}}}
&\textbf{Pre-edited}& {70.47\std{0.12}} & {18.77\std{0.09}} & {26.18\std{0.09}} & {24.12\std{0.17}} & {16.83\std{0.18}} & {28.56\std{0.18}}\\
&\textbf{FT-L}& {25.35\std{0.25}} & {14.21\std{0.12}} & {28.01\std{0.10}} & {23.57\std{0.16}} & {15.45\std{0.17}} & {29.41\std{0.18}}\\
&\textbf{ROME}& {77.56\std{0.17}} & {40.01\std{0.22}} & {27.35\std{0.09}} & {28.38\std{0.17}} & {22.31\std{0.19}} & {34.35\std{0.19}}\\
&\textbf{MEMIT}& {81.08\std{0.16}} & {51.43\std{0.26}} & {27.43\std{0.08}} & {30.13\std{0.18}} & {24.36\std{0.20}} & {36.96\std{0.20}}\\
&\textbf{UnKE}& \underline{97.25\std{0.07}} & {89.76\std{0.18}} & {26.97\std{0.08}} & {28.26\std{0.17}} & {21.18\std{0.19}} & {33.86\std{0.18}}\\
&\textbf{AnyEdit}& {97.13\std{0.07}} & \underline{90.78\std{0.18}} & \underline{29.97\std{0.08}} & \underline{34.59\std{0.19}} & \underline{32.27\std{0.22}} & \underline{45.38\std{0.20}}\\
 &\textbf{\method}& \textbf{99.23\std{0.04}} & \textbf{97.61\std{0.09}} & \textbf{56.23\std{0.15}} & \textbf{45.62\std{0.20}} & \textbf{41.89\std{0.22}} & \textbf{52.22\std{0.20}}\\

\midrule[1.0pt]

\multicolumn{8}{c}{\textbf{\benchmark-MQ}} \\
\midrule[1.0pt]

\multirow{7}{*}{\rotatebox{90}{{\textbf{Llama3-8B}}}} 
&\textbf{Pre-edited}& {69.67\std{0.13}} & {20.60\std{0.09}} & {27.74\std{0.09}} & {30.35\std{0.17}} & {25.88\std{0.21}} & {37.71\std{0.19}}\\
&\textbf{FT-L}& {22.65\std{0.18}} & {7.40\std{0.09}} & {28.04\std{0.10}} & {24.71\std{0.18}} & {18.16\std{0.19}} & {29.65\std{0.20}}\\
&\textbf{ROME}& {74.04\std{0.17}} & {36.10\std{0.20}} & {28.01\std{0.08}} & {31.37\std{0.17}} & {27.80\std{0.20}} & {40.47\std{0.19}}\\
&\textbf{MEMIT}& {78.86\std{0.19}} & {50.91\std{0.29}} & {29.22\std{0.08}} & {34.61\std{0.18}} & {31.94\std{0.23}} & {44.09\std{0.21}}\\
&\textbf{UnKE}& \underline{98.23\std{0.06}} & \underline{95.18\std{0.15}} & \underline{30.27\std{0.09}} & \underline{39.08\std{0.20}} & \underline{37.36\std{0.24}} & \underline{49.46\std{0.22}}\\
&\textbf{AnyEdit}& {97.53\std{0.07}} & {93.77\std{0.12}} & {29.96\std{0.08}} & {38.42\std{0.18}} & {36.62\std{0.22}} & {49.00\std{0.20}}\\
&\textbf{\method}& \textbf{98.71\std{0.07}} & \textbf{96.59\std{0.13}} & \textbf{66.99\std{0.14}} & \textbf{62.55\std{0.21}} & \textbf{65.86\std{0.23}} & \textbf{71.31\std{0.20}}\\

\midrule[1pt]
\midrule[1pt]

\multirow{7}{*}{\rotatebox{90}{{\textbf{Qwen2.5-7B}}}}
&\textbf{Pre-edited}& {69.08\std{0.12}} & {20.98\std{0.08}} & {28.28\std{0.09}} & {28.81\std{0.16}} & {22.94\std{0.19}} & {34.91\std{0.18}}\\
&\textbf{FT-L}& {23.03\std{0.22}} & {12.51\std{0.10}} & {27.27\std{0.09}} & {21.15\std{0.14}} & {14.29\std{0.16}} & {27.95\std{0.16}}\\
&\textbf{ROME}& {75.39\std{0.16}} & {40.29\std{0.19}} & {28.66\std{0.08}} & {31.73\std{0.17}} & {26.47\std{0.21}} & {39.39\std{0.19}}\\
&\textbf{MEMIT}& {79.44\std{0.16}} & {51.71\std{0.25}} & {28.54\std{0.08}} & {33.15\std{0.18}} & {26.75\std{0.20}} & {40.83\std{0.19}}\\
&\textbf{UnKE}& {94.88\std{0.11}} & {85.35\std{0.21}} & {29.31\std{0.09}} & {34.57\std{0.18}} & {29.81\std{0.21}} & {42.89\std{0.19}}\\
&\textbf{AnyEdit}& \underline{96.45\std{0.07}} & \underline{90.30\std{0.17}} & \underline{29.48\std{0.09}} & \underline{35.66\std{0.18}} & \underline{31.59\std{0.22}} & \underline{44.43\std{0.21}}\\
 &\textbf{\method}& \textbf{98.84\std{0.05}} & \textbf{96.98\std{0.10}} & \textbf{53.16\std{0.15}} & \textbf{44.01\std{0.20}} & \textbf{42.13\std{0.25}} & \textbf{51.10\std{0.22}}\\

\bottomrule[1.5pt]

\end{tabular}
}
\label{tab:main_results}
\end{table*}

\subsection{Experimental Setup}
\paragraph{Base LLMs.}
This study selects two categories of representative LLMs in this field: \textbf{Llama3-8B-Instruct}\footnote{\href{https://llama.meta.com/llama3}{https://llama.meta.com/llama3}} and \textbf{Qwen2.5-7B-Instruct}\cite{qwen2.5} as the base models.

\paragraph{Baseline Methods.}
For a comprehensive comparison, we select the following three representative categories of knowledge editing methods as baselines: (1) the fine-tuning-based method \textbf{FT-L}; (2) the classic "locate-then-edit" methods for structured editing scenarios, including \textbf{ROME}\cite{rome} and \textbf{MEMIT}\cite{memit}; (3) the "locate-then-edit" methods optimized for unstructured editing scenarios, including \textbf{AnyEdit}\cite{anyedit} and \textbf{UnKE}\cite{unke}. Details for each method are provided in Appendix~\ref{appendix:methods}.

\paragraph{Benchmark.}
We evaluate on \textbf{\benchmark}, our unstructured knowledge editing benchmark built on prior work for comprehensive assessment. Details are in Section~\ref{section:benchmark_construct}.

\paragraph{Implementation Details.} 
Following the experimental setup of works\cite{unke, anyedit}, this experiment sets the editing batch size to 1. More detailed experimental configurations and tools used can be found in Appendix~\ref{appendix:implementation_details}.

\subsection{Editing Performance Results}
\label{section:editing_performance_results}
Table~\ref{tab:main_results} reports the unstructured knowledge editing performance of different editing methods across various models and datasets. The experimental results show that {\method} significantly outperforms all baseline methods in both holistic generation quality and fine-grained knowledge recall.  

\textbf{(1) Holistic unstructured text generation performance is further improved.} In terms of the semantic similarity metric Bert-Score and the lexical overlap metric Rouge-L, {\method} achieves comprehensive superiority. Compared to the second-best baseline AnyEdit, {\method} improves Bert-Score and Rouge-L by an average of 1.98\% and 5.42\%, respectively. Notably, after editing the Llama3-8B model on the {\benchmark-UnKE} dataset, the improvement in Rouge-L reaches 
7.44\%.  

\textbf{(2) Fine-grained knowledge recall performance is significantly enhanced.} On sentence-level evaluation metrics (Bert-Score and Rouge-L), {\method} outperforms the second-best baseline AnyEdit by an average of 31.43\% and 17.74\%, respectively. On the fact-level metrics we further propose (HR and $C_{\text{LCS}}$), the corresponding average improvements are 20.07\% and 14.80\%.  

These results show that {\method} enhances both holistic text storage and fine-grained knowledge recall. Case studies are in Appendix~\ref{appendix:case_study}.

\begin{figure}[t]
    \centering
    \includegraphics[width=1\linewidth]{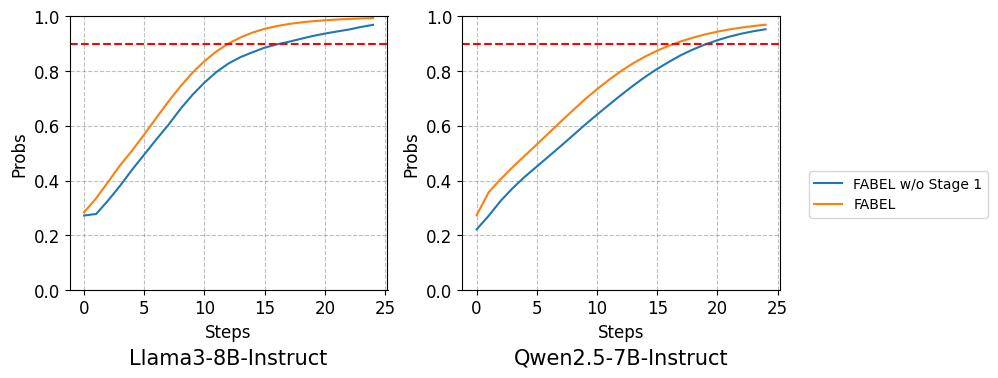}
    \caption{Performance comparison during optimization between FABEL (with Stage 1) and its ablated variant (without Stage 1) on the {\benchmark-UnKE}. The horizontal axis shows the number of optimization steps, and the vertical axis shows the average output probability of the target unstructured text.}
    \label{fig:prob_and_step_unke_results}
\end{figure}

\begin{figure}[t]
    \centering
    \includegraphics[width=1\linewidth]{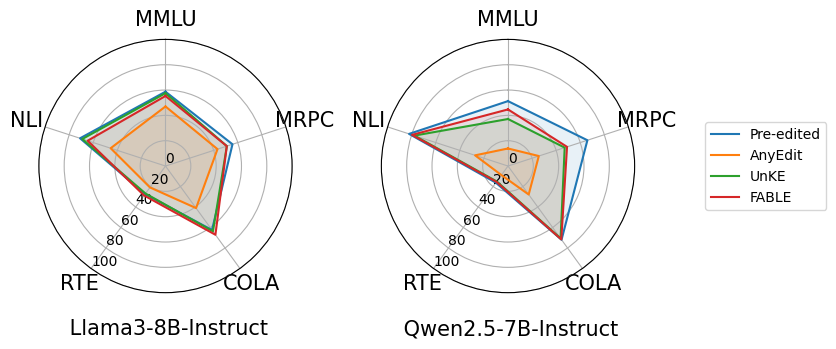}
    \caption{F1-scores of each task for unstructured editing methods on different models after editing on the {\benchmark-UnKE} dataset}
    \label{fig:general_unke_results}
\end{figure}

\begin{table*}[t]
\centering
\caption{The ablation experiment results of Llama3-8B-Instruct on the {\benchmark-UnKE} dataset, where "Layers" indicates the updated layers of $\mathcal{F}_{\text{fine}}$, "Number" denotes the quantity of fine-grained question-answer pairs injected in the first stage, "Method" refers to the ablation on the method itself, and "Augmentation" represents the ablation on data augmentation.}
\Large
\renewcommand{\arraystretch}{1} 
\resizebox{0.7\textwidth}{!}{%
\begin{tabular}{c|c|cc|cccc}
\toprule[1.5pt]

\multirow{2}{*}{\textbf{Ablation Dimensions}} & \multirow{2}{*}{\textbf{Configuration}} & \multicolumn{2}{c|}{\textbf{Holistic}} & \multicolumn{4}{c}{\textbf{Fine-grained}} \\ 
\cmidrule(lr){3-8}

& & \textbf{Bert-Score$\uparrow$} & \multicolumn{1}{c|}{\textbf{Rouge-L$\uparrow$}} &\textbf{Bert-Score$\uparrow$} & \textbf{Rouge-L$\uparrow$} & \textbf{HR$\uparrow$} & \textbf{\(C_{\text{LCS}}\)$\uparrow$}\\ 
\midrule[1.0pt]

\multirow{1}{*}{{\textbf{Base}}}
&\textbf{L=4,5,6; N=5 $\times$ S}& \textbf{99.36\std{0.04}} & \textbf{97.78\std{0.10}} & \textbf{65.63\std{0.16}} & \textbf{53.79\std{0.19}} & \textbf{53.31\std{0.23}} & \textbf{61.78\std{0.19}}\\

\midrule[1.0pt]
\multicolumn{8}{c}{\textbf{Ablation on Layers}} \\
\midrule[1.0pt]

\multirow{6}{*}{{\textbf{Layers}}}
&\textbf{L=3} & {99.51\std{0.04}} & {98.60\std{0.09}} & {63.18\std{0.16}} & {50.36\std{0.19}} & {48.33\std{0.23}} & {57.83\std{0.19}}\\
&\textbf{L=4} & {99.70\std{0.02}} & {98.79\std{0.08}} & {65.52\std{0.15}} & {52.93\std{0.19}} & {52.07\std{0.23}} & {61.09\std{0.19}}\\
&\textbf{L=5} & {99.68\std{0.02}} & {98.78\std{0.07}} & {64.93\std{0.15}} & {52.97\std{0.19}} & {51.88\std{0.24}} & {60.96\std{0.19}}\\
&\textbf{L=6} & {99.00\std{0.05}} & {96.66\std{0.12}} & {65.72\std{0.15}} & {53.96\std{0.19}} & {53.67\std{0.23}} & {61.90\std{0.19}}\\
&\textbf{L=4,5} & {99.67\std{0.03}} & {98.87\std{0.07}} & {65.92\std{0.15}} & {53.94\std{0.19}} & {53.17\std{0.23}} & {62.04\std{0.18}}\\
&\textbf{L=5,6} & {99.24\std{0.03}} & {96.80\std{0.12}} & {64.38\std{0.15}} & {52.72\std{0.19}} & {51.86\std{0.23}} & {60.41\std{0.19}}\\

\midrule[1.0pt]
\multicolumn{8}{c}{\textbf{Ablation on Number}} \\
\midrule[1.0pt]

\multirow{2}{*}{{\textbf{Number}}}
&\textbf{N=1 $\times$ S} & {99.65\std{0.02}} & {98.75\std{0.07}} & {57.97\std{0.16}} & {49.16\std{0.19}} & {46.90\std{0.23}} & {57.19\std{0.19}}\\
&\textbf{N=10 $\times$ S} & {97.11\std{0.09}} & {92.23\std{0.19}} & {66.42\std{0.15}} & {53.69\std{0.20}} & {53.79\std{0.23}} & {61.75\std{0.19}}\\

\midrule[1.0pt]
\multicolumn{8}{c}{\textbf{Ablation on Method}} \\
\midrule[1.0pt]
\multirow{2}{*}{{\textbf{Method}}}
&\textbf{w/o Stage1} & {98.63\std{0.05}} & {94.92\std{0.15}}  & {23.29\std{0.08}}  & {14.98\std{0.11}}  & {7.56\std{0.13}}  & {20.42\std{0.14}} \\
&\textbf{w/o Stage2} & {45.98\std{0.20}} & {9.13\std{0.08}}  & {72.00\std{0.14}}  & {55.36\std{0.19}}  & {55.04\std{0.24}}  & {62.47\std{0.19}} \\

\midrule[1.0pt]
\multicolumn{8}{c}{\textbf{Ablation on Augmentation}} \\
\midrule[1.0pt]
\multirow{2}{*}{{\textbf{Augmentation}}}
&\textbf{UnKE+Augmentation} & {97.17\std{0.10}} & {92.12\std{0.21}} & {68.54\std{0.15}} & {54.26\std{0.20}} & {54.11\std{0.23}} & {61.77\std{0.19}}\\
&\textbf{AnyEdit+Augmentation} & {61.57\std{0.23}} & {17.71\std{0.18}} & {61.50\std{0.15}} & {44.63\std{0.19}} & {41.31\std{0.22}} & {51.00\std{0.19}}\\

\bottomrule[1.5pt]

\end{tabular}
}
\label{tab:ablation_results}
\end{table*}

\subsection{Analysis of the Performance Enhancement}
Based on the findings in Section~\ref{section:editing_performance_results}—that {\method}, compared to baseline methods like UnKE and AnyEdit, not only improves fine-grained knowledge recall but also further enhances the overall performance of generating unstructured text—we delve into the underlying reasons in this section. In stage two (semantic integration), {\method} optimizes the residual vector $\delta_h$ through gradient descent to search for the optimal target semantic key $k^*_h$, thereby maximizing the probability of the model generating the target unstructured text. To reveal this process, we take the model edited on the {\benchmark-UnKE} dataset as an example and plot the curve of the model’s average output probability as a function of optimization steps (Figure~\ref{fig:prob_and_step_unke_results}). The figure compares the optimization trajectories of the full two-stage method (FABEL) and the method using only stage two (FABEL w/o Stage1), from which two key observations can be made:

\textbf{(1) The introduction of stage one significantly improves the initial probability.} As shown in the figure, FABLE (orange) starts higher than FABLE w/o Stage One (blue) on both Llama3-8B-Instruct and Qwen2.5-7B-Instruct.  Therefore, higher initial probability facilitates editing and improves success \cite{anyedit}, explaining the enhanced final text generation of {\method}.

\textbf{(2) Stage one effectively reduces the difficulty of editing and accelerates convergence.} The red dashed line marks a 0.9 probability threshold. FABLE with Stage One reaches it faster—e.g., in approximately 10 steps on Llama3-8B-Instruct versus in approximately 15 without Stage One—showing that prior atomic fact injection simplifies subsequent semantic integration.

These results show that {\method}'s hierarchical key-value architecture and two-stage editing improve both unstructured knowledge modeling and editing efficiency. Additional datasets are in Appendix~\ref{appendix:more_analysis}.

\subsection{General Capability Evaluation}
Unstructured editing methods have demonstrated promising editing performance in Section~\ref{section:editing_performance_results}. However, their impact on the general capabilities of the models remains unclear. To address this, we selected several unstructured editing methods with strong editing performance—UnKE, AnyEdit, and {\method}—to further examine their effects on general capabilities. Previous work \cite{unke} employed a relatively narrow evaluation of the general capabilities of edited models, testing solely on the MMLU dataset \cite{mmmlu}. To assess the impact of editing methods on model general capabilities more comprehensively, we introduced an additional 5 datasets involving language understanding and logical reasoning for a multi-faceted evaluation; specific details are provided in Appendix~\ref{appendix:details_general_evaluation}.

Taking the models edited on the {\benchmark-UnKE} dataset as an example, the results of their general capability assessment are shown in Figure~\ref{fig:general_unke_results}. The key findings are summarized as follows:
\textbf{(1) Baseline methods exhibit varying degrees of degradation in general capabilities}: For models edited using the AnyEdit method, performance significantly declined across multiple tasks. While the UnKE method outperformed AnyEdit, it still showed a noticeable drop in performance on the Qwen2.5-7B-Instruct model. This indicates that baseline methods still have shortcomings in preserving the stability of the original knowledge unrelated to the edits.
\textbf{(2) {\method} demonstrates an outstanding ability to preserve general capabilities}: 
On Llama3-8B-Instruct, {\method} matches UnKE in maintaining pre-editing performance. On Qwen2.5-7B-Instruct, although all methods decline slightly, {\method} best preserves general capabilities. This shows it updates unstructured knowledge accurately while minimizing interference. Additional results are in Appendix~\ref{appendix:more_general_results}.

\subsection{Ablation Study}
\label{section:ablation_study}
Taking the Llama3-8B-Instruct model as an example, we conduct editing tasks on the {\benchmark-UnKE} dataset and perform ablation experiments from multiple dimensions. The results are shown in Table~\ref{tab:ablation_results}. Specifically, we analyze the following four aspects: \textbf{(1) Layer Selection}: Updating the 4th, 5th, and 6th layers in $\mathcal{F}_{\text{fine}}$ simultaneously achieves the best trade-off between overall performance and fine-grained performance, validating the effectiveness of this combination. \textbf{(2) Data Volume}: The number of fine-grained question-answer pairs has an optimal value. Insufficient quantity (N=1×S) leads to inadequate fine-grained performance, while excessive quantity (N=10×S) significantly impairs overall performance (Rouge-L drops to 92.23) when fine-grained metrics saturate. Experiments indicate that N=5×S is the ideal setting. \textbf{(3) Method Necessity}: Removing the first stage (w/o Stage1) or second stage (w/o Stage2) results in declines in both overall and fine-grained performance, demonstrating that fine-grained knowledge injection in the first stage and holistic semantic integration in the second stage serve as the critical foundation. \textbf{(4) Data Augmentation Comparison}: To investigate whether performance improvement is merely due to data augmentation, we augment baseline methods with the same fine-grained question-answer pairs. For UnKE, results show that while fine-grained metrics improve slightly, overall performance is significantly impaired (Rouge-L: 92.12), and the combined performance is lower than that of the proposed method. Consistent with the UnKE, augmenting AnyEdit with our data leads to a decline in its holistic performance and fails to achieve the balanced effectiveness exhibited by {\method}. This indicates that performance gains primarily stem from the {\method} method itself, rather than simple data augmentation.

\section{Related Work}
Model editing has become a rapidly growing research area, as it enables LLMs to update internal knowledge without full retraining~\cite{ke,mend,malmen,instructedit,serac,calinet,t-patcher,grace,melo,wise,rect,prune,alphaedit,DBLP:journals/corr/abs-2505-15702}. Locate-then-edit is a dominant paradigm, enabling precise updates to parameters associated with triple-based knowledge. Representative methods include KN~\cite{kn}, ROME~\cite{rome}, and MEMIT~\cite{memit}. Recent efforts extend editing to unstructured knowledge, evaluated on benchmarks like AKEW~\cite{akew} and UnKEBench~\cite{unke}. Methods such as UnKE~\cite{unke}, AnyEdit~\cite{anyedit}, and $\mu$KE~\cite{DBLP:journals/corr/abs-2504-01196} adapt locate-then-edit for noisy or lengthy content, while DEM~\cite{DBLP:conf/emnlp/HuangW0024} focuses on efficient parameter localization.

\section{Conclusion}
In this paper, we address the limitation that existing unstructured model editing methods focus on holistic text recall but lack reliable fine-grained fact access. We propose {\method}, a hierarchical framework that first anchors fine-grained facts in shallow layers and then updates deeper layers for coherent surface-form generation. We also introduce {\benchmark} for evaluating fine-grained fact recall. Experiments demonstrate that {\method} improves factual accuracy while maintaining overall text quality. 

\section*{Statements}

\paragraph{LLM Usage.} LLMs were used only for sentence polishing and grammar correction, not for content creation or scientific claims.

\paragraph{Ethics statements.} All code and datasets used in this work are publicly available. While LLMs offer substantial benefits, they can be misused to generate harmful content, misinformation, or offensive material. Our research on unstructured knowledge editing is conducted ethically and solely for scientific purposes, aiming to improve LLM reliability, interpretability, and factual accuracy. The methods are not intended for malicious use or privacy violations, and we strongly encourage rigorous validation and oversight to ensure responsible application.

\section*{Limitation}
While our method significantly enhances fine-grained fact recall while maintaining state-of-the-art holistic editing performance, we observe a slight degradation in the generalization of surface-form construction. This may stem from the constrained preservation of fine-grained knowledge in Stage Two, highlighting an inherent trade-off between achieving coherent holistic generation and ensuring precise fine-grained fact access. Future work will explore strategies to further balance holistic fluency and fine-grained factual accuracy.

The present experiments and evaluations are conducted exclusively under a single-edit scenario (batch size = 1). While this setup suffices to demonstrate the feasibility of unstructured-knowledge updates, it omits the more realistic and challenging regimes of batched and sequential editing. Specifically, two limitations are evident: (1) the study does not address batch editing, where multiple independent knowledge-update requests are processed simultaneously—a common real-world situation that heightens the risk of internal knowledge conflicts and parameter interference; and (2) it does not examine sequential editing, in which the model undergoes a series of potentially interrelated updates over time, demanding long-term consistency and robust mitigation of catastrophic forgetting. Future work should extend along both of these dimensions.

\section*{Acknowledgments}
This work is supported by the National Natural Science Foundation of China (No.U24A20335).


\bibliography{custom}

\appendix

\section{Benchmark}
\subsection{Datasets Construction}
\label{appendix:datasets_construction}
\paragraph{Generation of Fine-Grained QA Pairs}
We employ the GPT-4o\cite{gpt-4o} model to generate five corresponding fine-grained question-answer (QA) pairs for each target output \(A\) in AKEW (CounterFact) and AKEW (MQuAKE). The generation process adheres to the following principles:
\begin{itemize}
    \item \textbf{Diverse Perspectives:} Questions are formulated from different angles using varied interrogatives (e.g., what, when, how) to ensure coverage of distinct aspects of the target output, with non-repetitive answer content.
    \item \textbf{Entity Consistency:} All entities mentioned in the answers must originate from the original target output \(A\).
    \item \textbf{Length Control:} Each answer is constrained to a maximum of 15 tokens.
    \item \textbf{Clarity of Expression:} Both questions and answers must be concise, grammatically correct, and avoid meta-references to the text's own structure.
\end{itemize}
To ensure quality, we verify via string matching that all answer content is derived from the target output \(A\). The specific prompt template is provided in Table~\ref{template:generate_fine_grained_qa}.

\paragraph{Key Knowledge Phrase Extraction}
For each dataset, we similarly utilize GPT-4o in a few-shot manner, instructing the model to perform the following operations for each answer:
\begin{itemize}
    \item \textbf{Atomize:} Decompose the answer into its minimal factual units.
    \item \textbf{Concise:} Remove modifiers, fillers, and redundant connecting words.
    \item \textbf{Stay-source:} Ensure the extracted phrases are strictly sub-sequences of the original answer text.
    \item \textbf{Split:} If an answer contains multiple independent facts, split them into separate, parallel phrases.
    \item \textbf{Objectify:} Retain only objective factual statements, filtering out subjective or evaluative language.
\end{itemize}
The extraction results are also validated via string matching against the original answers to ensure faithfulness. The corresponding prompt template is shown in Table~\ref{template:key_knowledge_phrase_extraction}.

\subsection{Evaluation Metrics}
\label{appendix:evaluation_metrics}
The evaluation in this study employs the following primary metrics:
\paragraph{Lexical Similarity.}
We use ROUGE-L to measure the n-gram overlap between the model-generated text and the target answer. This metric assesses the surface-level accuracy of the generated content.
\paragraph{Semantic Similarity.}
To complement the limitations of lexical metrics, we compute the semantic similarity Bert-Score using the all-MiniLM-L6-v2 encoder. This evaluates whether the model has genuinely understood the textual meaning rather than merely repeating surface-level patterns.
\paragraph{Hit Rate (HR).}
This metric measures the model's ability to accurately recall key knowledge. Formally, for a fine-grained gold answer \(A^{fine}\), let the set of extracted key phrases be \(KP = \{kp_1, kp_2, \dots, kp_m\}\), and the model output be \(O^{fine}\). An indicator function \(\mathbb{I}(kp_i, O)\) is defined for each key phrase:

\begin{equation}
\mathbb{I}(kp_i, O) =
\begin{cases}
1, & \text{if } kp_i \text{ is a substring of } O^{fine}, \\
0, & \text{otherwise}.
\end{cases}
\end{equation}

The Hit Rate is then the average of these indicators over all key phrases:

\begin{equation}
HR = \frac{1}{m} \sum_{i=1}^{m} \mathbb{I}(kp_i, O^{fine}).
\end{equation}

\paragraph{Longest Common Subsequence Coverage (\(C_{\text{LCS}}\)).}
This metric quantifies the completeness of the model output in covering the content of the answer. Let the fine-grained gold answer \(A^{fine}\) be tokenized into a word sequence \(A^{fine}= [a^{fine}_1, a^{fine}_2, \dots, a^{fine}_n]\) and the model output \(O^{fine}\) into \(O^{fine} = [o^{fine}_1, o^{fine}_2, \dots, o^{fine}_l]\). Denoting the length (in words) of the longest common subsequence between \(A^{fine}\) and \(O^{fine}\) as \(\text{LCS}(A^{fine}, O^{fine})\), the coverage \(C_{\text{LCS}}\) is defined as:

\begin{equation}
C_{\text{LCS}} = \frac{\text{LCS}(A^{fine}, O^{fine})}{n}
\end{equation}

where \(n\) is the total number of words in \(A^{fine}\). The LCS length is computed using a standard dynamic programming algorithm. Let \(\text{LCS}(i, j)\) represent the length of the \(\text{LCS}\) between the prefixes \(A^{fine}[1:i]\) and \(O^{fine}[1:j]\). Let \(\text{LCS}(i,0) = 0\) and \(\text{LCS}(0,j) = 0\), For \(\forall i,j \geq 0\), If \(a^{fine}_i = o^{fine}_j\), \(\text{LCS}(i,j) = \text{LCS}(i-1,j-1)+1\), else \(\text{LCS}(i,j) = \max\bigl\{\text{LCS}(i-1,j),\,\text{LCS}(i,j-1)\bigr\}\). Finally, \(\text{LCS}(A^{fine}, O^{fine}) = \text{LCS}(n, l)\).

\begin{table*}[ht]
\centering
\caption{Prompt Template for Generating Fine-Grained QA Pairs}
\label{template:generate_fine_grained_qa}
\resizebox{2.0 \columnwidth}{!}{
\begin{tabular}{p{1.3\linewidth}}
\hline
\textbf{Prompt Template for Generating Fine-Grained QA Pairs}  \\
\hline
You are asked to generate some short question-answer pairs based on the specified <Text>. These question-answer pairs mainly ask questions about the knowledge entities in the <Text>, and the answers should be the knowledge entities being asked. \\
\\
Rules: \\
- Diversify Perspectives: Formulate questions from multiple angles to minimize overlapping answers. Use diverse question words (What, When, Where, Why, How, etc.); avoid "Who" or "Whose". Avoid yes/no questions unless they are highly informative. \\
- Entity Inclusion: Ensure all entities mentioned in the answers are present in the <Text>. \\
- Length Constraint: Each answer must not exceed 15 tokens. \\
- Fluency and Clarity: Ensure that both questions and answers are clear, grammatical, and concise. Do not include meta-phrases such as “in the sentence”, “according to the sentence”, “mentioned in the text”, etc. inside the questions themselves; the question must read as if it were asked in real life, not about the text. \\
- The <Output> formats the pairs as a JSON object with "questions" and "answers" lists. \\
\{ \\
~~~~"questions": [] \\
~~~~"answers": [] \\
\} \\
\\
Now generate 5 question-answer pairs for the following <Text>. \\
<Text>: \{text\} \\
<Output>: \\
\hline
\end{tabular}
}
\end{table*}

\begin{table*}[ht]
\centering
\caption{Prompt Template for Key Knowledge Phrase Extraction}
\label{template:key_knowledge_phrase_extraction}
\resizebox{2.0 \columnwidth}{!}{
\begin{tabular}{p{1.3\linewidth}}
\hline
\textbf{Prompt Template for Key Knowledge Phrase Extraction}  \\
\hline
You are an assistant designed to extract minimal factual answers from any given <Answer>.\\
\\
Rules:\\
- Break down the <Answer> into the smallest possible factual units. Each unit should represent a standalone piece of information.\\
- Remove any stylistic flourishes, adjectives, adverbs, or unnecessary connectors (e.g., "over" in "over a decade" is retained only if it's part of the factual unit, as shown in the example).\\
- All tokens of the extracted answer must strictly reside within the <Answer>.\\
- If the <Answer> contains multiple distinct facts separated by conjunctions like "and" or commas, split them into separate items in the list.\\
- Focus solely on factual content; ignore any subjective or embellished language.\\
\\
Example:\\
<Question>: How long has George Rankin been involved in politics?\\
<Answer>: Over a decade.\\
<Output>:\\
\{\\
~~~~"answers": ["over a decade"]\\
\}\\
\\
<Question>: What positions has George Rankin held in politics?\\
<Answer>: City council member and state representative.\\
<Output>:\\
\{\\
  "answers": ["city council member", "state representative"]\\
\}\\
\\
<Question>: Where is George Rankin frequently quoted?\\
<Answer>: Local and national news outlets.\\
<Output>:\\
\{\\
~~~~"answers": ["local news outlets", "national news outlets"]\\
\}\\
\\
<Question>: What did John Mayne discuss in his interview with The Huffington Post?\\
<Answer>: His passion for journalism and his commitment to reporting on important issues.\\
<Output>:\\
\{\\
~~~~"answers": ["passion for journalism", "commitment to reporting on important issues"]\\
\}\\
\\
Now generate the minimal factual answers for the following input.\\
<Question>: \{question\}\\
<Answer>: \{answer\}\\
<Output>:\\
\hline
\end{tabular}
}
\end{table*}

\section{Multi-Aspect Fine-Grained Knowledge Extraction for Unstructured Text}
\label{appendix:multi-aspect-extraction}
In target unstructured texts, there often exist various types of fine-grained knowledge. To extract this knowledge as comprehensively and non-overlap as possible from different perspectives, we designed the following pipeline: 

First, the target text is segmented into several sub-sentences using the NLTK toolkit. 

Next, a word-count threshold is set to filter out sub-sentences shorter than the threshold, thereby reducing noise caused by insufficient information. 

Then, for each retained sub-sentence, multiple high-quality natural-language questions are generated from various angles based on GPT-4o (details of the templates are shown in Table~\ref{template:multi-aspect-extraction}). The question generation follows these principles: (1) Answerability: the question should be directly answerable based on the sub-sentence, with the answer either explicitly stated or reasonably inferable from it; (2) Clarity: the question must be grammatically correct, clearly expressed, highly consistent with the sub-sentence content, and unambiguous; (3) Non-binary: yes/no questions are avoided unless their answers carry high informational value, and open-ended interrogatives are encouraged; (4) Diversity: varied interrogative words are used to prevent repetitive questioning patterns.

Subsequently, for each generated question, GPT-4o is called once more to produce a corresponding answer based on the target text, forming an initial question-answer pair. The question-answer pairs corresponding to one sub-sentence constitute a cluster. We first filter out those pairs whose answers are primarily derived from that sub-sentence, and then perform merging and deduplication: if the answer of one pair completely contains the content of another answer, they are regarded as more complete expressions of the same questioning perspective; the former is retained and the latter is removed. After this step, each sub-sentence ultimately retains a set of the most diverse and information-complete fine-grained question-answer pairs.

Finally, to increase the number of retained questions, each kept question is used as a seed question, and GPT-4o is employed to generate multiple semantically similar, fine-grained questions that point to the same answer, thereby further expanding the set of fine-grained knowledge questions.

\begin{table*}[ht]
\centering
\caption{Prompt Template for Multi-Aspect Fine-Grained Knowledge Extraction}
\label{template:multi-aspect-extraction}
\resizebox{2.0 \columnwidth}{!}{
\begin{tabular}{p{1.3\linewidth}}
\hline
\textbf{Prompt Template for Multi-Aspect Fine-Grained Knowledge Extraction}  \\
\hline
You are a question generation expert. Your task is to generate **non-overlapping** natural language questions based on the given sentence. \\
\\
Rules:\\
- Each question must be answerable directly using only the information in the sentence. The answer to each question must appear in the sentence (verbatim or semantically contained)\\
- The question should be clear, grammatical, and relevant.\\
- Avoid yes/no questions unless they are highly informative.\\
- Use diverse question words (What, When, Where, Why, How, etc.); avoid "Who" or "Whose".\\
- Avoid pronouns; use specific names, places, etc.\\
- Do not include meta-phrases such as “in the sentence”, “according to the sentence”, “mentioned in the text”, etc. inside the questions themselves; the question must read as if it were asked in real life, not about the text.\\
- Do not repeat the same information in two different questions; ensure every answer is mutually exclusive in content. \\
\\
Example:\\
Paragraph: "Marie Curie discovered radium in 1898 with her husband Pierre Curie. This breakthrough came after years of painstaking work in a leaky, unheated shed, where the Curies processed tons of pitchblende residue to isolate minute quantities of the glowing new element. Marie herself coined the name "radium" from the Latin word for "ray", captivated by its mysterious, persistent luminescence.\\
Sentence: "Marie Curie discovered radium in 1898 with her husband Pierre Curie."\\
Output:\\
\{\\
  "questions": ["Who discovered radium in 1898?", "What element did Marie Curie discover?", "When did Marie Curie and Pierre Curie discover radium?"]\\
\}\\
\\
Now generate questions for the following sentence (no fixed count; stop when angles are exhausted):\\
\\
Paragraph: \{paragraph\}\\
Sentence: "\{sentence\}"\\
Output:\\
\hline
\end{tabular}
}
\end{table*}

\section{Experimental Setup}
\label{appendix:setups}
\subsection{Baseline Methods}
\label{appendix:methods}
Here, we outline the baseline knowledge editing methods employed for comparison in this study:
\paragraph{FT-L.} The FT-L method adjusts designated layers of the LLM via autoregressive loss minimization to incorporate new knowledge.

\paragraph{ROME.} ROME is a model editing technique that localizes and updates factual associations in autoregressive transformers by identifying critical mid-layer MLP modules as key-value memories. It computes a subject-specific key from hidden states and optimizes a value vector to represent new knowledge, then applies a rank-one weight update to the MLP projections, effectively inserting facts while maintaining generalization and specificity.

\paragraph{MEMIT.} MEMIT, building upon ROME, is a scalable multi-layer editing algorithm. It efficiently integrates large-scale factual updates into LLMs by computing explicit parameter adjustments, thereby modifying memories while preserving the model's overall integrity.

\paragraph{AnyEdit.} AnyEdit is an autoregressive editing paradigm designed to overcome the limitations of single-token editing methods in large language models. It breaks down long-form knowledge into sequential chunks and iteratively edits the key token in each chunk, leveraging the Chain Rule of Mutual Information to ensure consistent and accurate generation of diverse formats.

\paragraph{UnKE.} UnKE is a method designed to edit unstructured knowledge in large language models by extending previous approaches in two key dimensions: it replaces local layer key-value storage with non-local block key-value storage to better represent distributed knowledge across layers, and employs cause-driven optimization that edits the last token directly to preserve context without needing term localization, thereby handling complex, long-form unstructured data effectively.

\subsection{Implementation Details}
\label{appendix:implementation_details}
All experiments were conducted on a single NVIDIA A100 (80 GB) GPU. To control variables and ensure a fair comparison, all methods in this study were implemented based on the widely-used model editing toolkit EasyEdit\cite{easyedit}, with its default optimal hyperparameters adopted.

\subsection{Time Cost}
\label{appendix:time_cost}
Table~\ref{tab:time_cost} further compares the editing efficiency of different methods. It can be observed that while FT-L is the fastest, it sacrifices editing effectiveness (see Table~\ref{tab:main_results}), revealing an inherent efficiency–effectiveness trade-off in this task. The relatively higher time cost of {\method} is a direct consequence of its design objective: to achieve more reliable unstructured knowledge editing with minimal side effects, the method employs a refined multi-step optimization process. Although this design increases the per-edit computational cost, it yields significant advantages in terms of editing performance and the preservation of general capabilities. Improving time efficiency will be a clear direction for future optimization.

\begin{table}[t]
\centering
\caption{\footnotesize Time cost of different methods across various models and datasets.}
\large
\renewcommand{\arraystretch}{0.5}
\resizebox{0.5\textwidth}{!}{
\begin{tabular}{c|c|ccc}
\toprule[1.5pt]
\textbf{Model} & \textbf{Method} & \textbf{{\benchmark-UnKE}(/s)} & \textbf{{\benchmark-CF}(/s)} & \textbf{{\benchmark-MQ}(/s)} \\
\midrule
\multirow{6}{*}{\rotatebox{90}{{LLaMA3-8B}}} 
 & FT-L & 3.76 & 3.44 & 3.55 \\
 & ROME & 17.74 & 15.02 & 16.78 \\
 & MEMIT & 54.50 & 46.43 & 39.28 \\
 & UnKE & 33.79 & 29.84 & 30.54 \\
 & AnyEdit & 20.99 & 15.55 & 15.94 \\
 & {\method} & 356.71 & 340.18 & 329.63 \\
\midrule
\multirow{6}{*}{\rotatebox{90}{{Qwen2.5-7B}}} 
 & FT-L & 2.90 & 2.74 & 2.73 \\
 & ROME & 72.44 & 60.37 & 61.53 \\
 & MEMIT & 54.95 & 45.96 & 46.78 \\
 & UnKE & 33.25 & 32.04 & 31.65 \\
 & AnyEdit & 22.83 & 21.64 & 21.73 \\
 & {\method} & 330.04 & 322.12 & 309.77 \\
\bottomrule[1.5pt]
\end{tabular}
}
\label{tab:time_cost}
\end{table}

\subsection{Details of General Capability Evaluation Datasets}
\label{appendix:details_general_evaluation}
To assess the impact of knowledge editing on the model’s general capabilities, we selected five datasets covering different linguistic understanding and reasoning tasks. For each editing sample, we uniformly sampled 5 data instances per task category (25 instances in total) to form an evaluation subset, and used the F1-score to measure the model’s post-editing performance. Each dataset is described in detail below:
\paragraph{CoLA.} CoLA \cite{cola} evaluates grammatical acceptability via binary classification of single-sentence judgments.

\paragraph{MMLU.} MMLU \cite{mmmlu} measures multi-task accuracy across diverse domains, focusing specifically on zero-shot and few-shot learning scenarios in text models.

\paragraph{NLI.} NLI \cite{nli} assesses language understanding by requiring models to identify logical relationships—such as entailment, contradiction, or neutrality—between pairs of sentences.

\paragraph{MRPC.} MRPC \cite{mrpc} evaluates semantic equivalence detection, where models determine whether two sentences convey the same meaning.

\paragraph{RTE.} RTE \citep{rte} examines whether a premise sentence logically supports a given hypothesis.

\section{More Experimental Results}
\subsection{More Analysis of the Performance Enhancement Results}
\label{appendix:more_analysis}
\begin{figure}[t]
    \centering
    \includegraphics[width=1\linewidth]{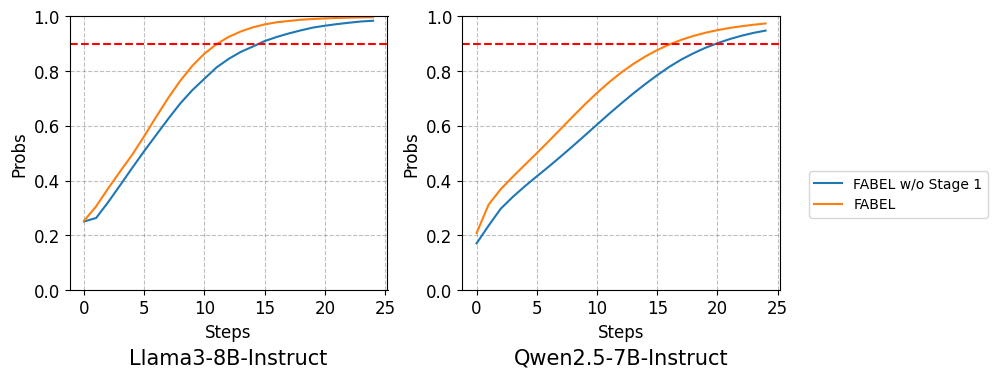}
    \caption{Performance comparison during optimization between FABEL (with Stage 1) and its ablated variant (without Stage 1) on the {\benchmark-UnKE}. The horizontal axis shows the number of optimization steps, and the vertical axis shows the average output probability of the target unstructured text.}
    \label{fig:prob_and_step_cf_results}
\end{figure}

\begin{figure}[t]
    \centering
    \includegraphics[width=1\linewidth]{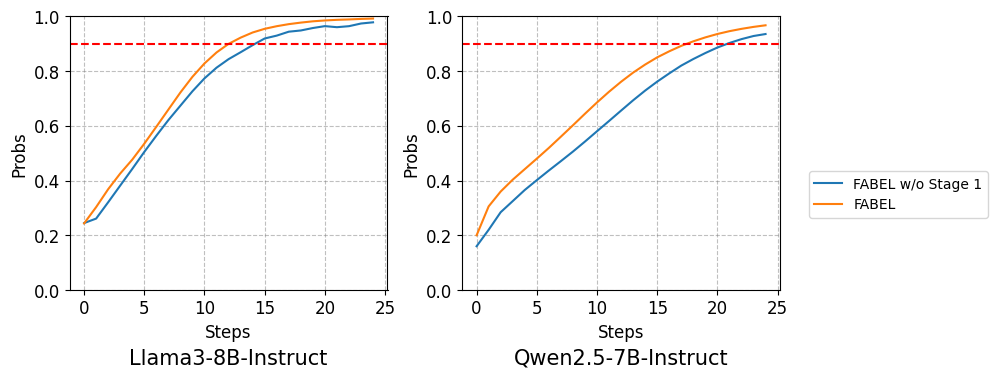}
    \caption{Performance comparison during optimization between FABEL (with Stage 1) and its ablated variant (without Stage 1) on the {\benchmark-UnKE}. The horizontal axis shows the number of optimization steps, and the vertical axis shows the average output probability of the target unstructured text.}
    \label{fig:prob_and_step_mquake_results}
\end{figure}

Figure~\ref{fig:prob_and_step_cf_results} and Figure~\ref{fig:prob_and_step_mquake_results} illustrate the optimization process of the proposed method on the {\benchmark-CF} and {\benchmark-MQ} datasets, respectively. As shown in the figures, the complete method with stage one (injection of atomic facts), i.e., {\method}, demonstrates both higher initial probability and faster convergence rate across the two different scenarios, consistent with the observations on the {\benchmark-UnKE} dataset. This further validates the effectiveness and robustness of {\method} across diverse tasks and settings, highlighting its strong generalizability.

\subsection{More General Capability Evaluation Results}
\label{appendix:more_general_results}
\begin{figure}[t]
    \centering
    \includegraphics[width=1\linewidth]{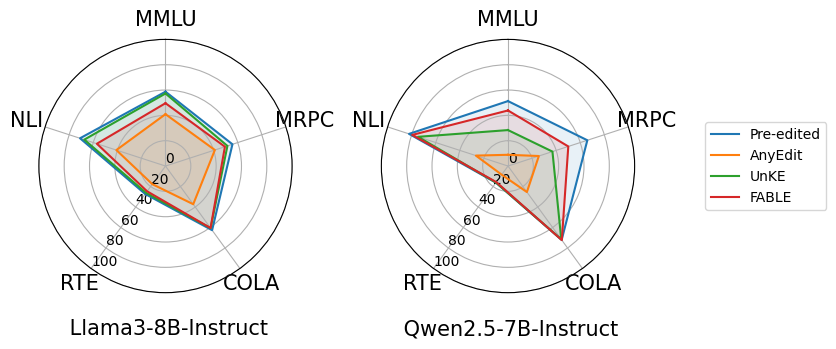}
    \caption{F1-scores of each task for unstructured editing methods on different models after editing on the {\benchmark-CF} dataset}
    \label{fig:general_cf_results}
\end{figure}

\begin{figure}[t]
    \centering
    \includegraphics[width=1\linewidth]{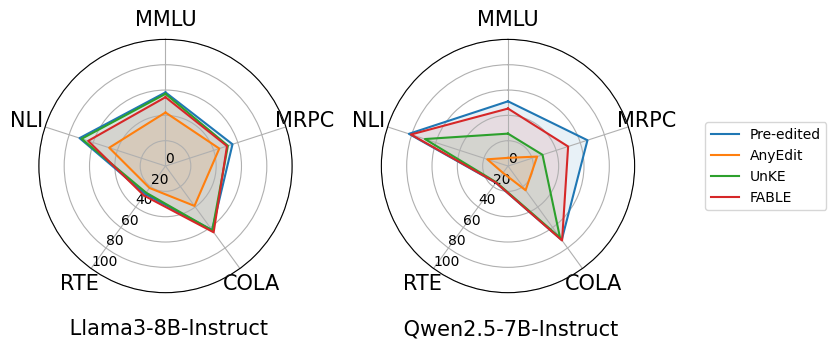}
    \caption{F1-scores of each task for unstructured editing methods on different models after editing on the {\benchmark-MQ} dataset}
    \label{fig:general_mquake_results}
\end{figure}

To further evaluate the performance of editing methods under the influence of different editing datasets, we additionally present the general capabilities of models after editing on the {\benchmark-CF} and {\benchmark-MQ} datasets, with the results shown in Figure~\ref{fig:general_cf_results} and Figure~\ref{fig:general_mquake_results}, respectively. Overall, {\method} still demonstrates relatively more stable retention of general capabilities across the evaluated tasks. However, its specific performance is influenced by factors such as model architectures and types of editing datasets, leading to variations under different experimental settings.

\subsection{Conflicting Unstructured Knowledge Updates}
The "conflicting unstructured knowledge updates" means the model must learn two different descriptions about the same subject simultaneously—is indeed a crucial test for evaluating the robustness and practicality of an editing method. We have given this serious consideration and conducted preliminary explorations. We manually constructed a small-scale Conflicting Editing DataSet (100 samples) based on the existing {\benchmark}-UnKE dataset. Each sample contains two different unstructured descriptions about the same subject, simulating real-world knowledge conflicts. We conducted experiments using the Llama3-8B-Instruct model and the {\method} method. The preliminary results are shown in the Table~\ref{tab:conflicting}.

The experimental results show that in the conflicting scenario, all performance metrics of {\method} experience a slight decline compared to the standard non-conflicting scenario. This is expected, indicating that contradictory knowledge indeed poses an additional challenge to the model's parameter updates and knowledge retention. However, it is noteworthy that the magnitude of the performance drop is relatively limited. Even under the conflicting setting, {\method} maintains strong capabilities in both holistic text generation (Rouge-L: 93.04) and fine-grained fact recall (HR: 49.62). This preliminary finding suggests that the hierarchical fact-anchoring strategy employed by {\method} provides a certain buffer against the conflicts arising from direct parameter overwriting, demonstrating potential robustness when facing contradictory knowledge updates.

\begin{table*}[t]
\centering
\caption{The experiment results of Llama3-8B-Instruct on the {\benchmark-UnKE} dataset and Conflicting Editing DataSet.}
\Large
\renewcommand{\arraystretch}{1} 
\resizebox{0.7\textwidth}{!}{%
\begin{tabular}{c|cc|cccc}
\toprule[1.5pt]

\multirow{2}{*}{\textbf{Mehtod}} & \multicolumn{2}{c|}{\textbf{Holistic}} & \multicolumn{4}{c}{\textbf{Fine-grained}} \\ 
\cmidrule(lr){2-7}

& \textbf{Bert-Score$\uparrow$} & \multicolumn{1}{c|}{\textbf{Rouge-L$\uparrow$}} &\textbf{Bert-Score$\uparrow$} & \textbf{Rouge-L$\uparrow$} & \textbf{HR$\uparrow$} & \textbf{\(C_{\text{LCS}}\)$\uparrow$}\\ 
\midrule[1.0pt]
\textbf{{\method} (Non-conflicting)} & {99.36\std{0.04}} & {97.78\std{0.10}} & {65.63\std{0.16}} & {53.79\std{0.19}} & {53.31\std{0.23}} & {61.78\std{0.19}}\\
\textbf{{\method} (Conflicting)} & {98.91\std{0.03}} & {93.04\std{0.15}} & {60.33\std{0.09}} & {50.48\std{0.11}} & {49.62\std{0.14}} & {56.39\std{0.08}}\\
\bottomrule[1.5pt]

\end{tabular}
}
\label{tab:conflicting}
\end{table*}

\subsection{Representation Comparison}
In unstructured editing methods (e.g., UnKE, AnyEdit), the standard optimization target is typically the hidden state of the last token of the complete input sequence, not a specific token representing the "subject word" within the fact. This has become an empirical setup adopted by many works in this field. As an advancement within the unstructured editing framework, our method follows this setup in its first stage. This aims to maintain comparability with baseline methods while exploring a superior hierarchical decoupling design. We acknowledge that in earlier structured editing approaches, the last token of the subject word (not the last token of the sequence) is often used as the clean fact representation. While we recognize this might lead to contextual mixing, it also makes the representation more suitable for the subsequent unstructured text generation objective.

In order to empirically evaluate the advantages and disadvantages of using the "last sequence token" versus the "last subject word token" as the fact representation, the experimental results are presented as Table~\ref{tab:representation}. The experiments above take editing the Llama3-8B-Instruct model on the {\benchmark}-UnKE dataset as an example.

The experimental results indicate that all metrics for {\method} (Last Subject Word Token) are lower than those of our proposed {\method} (Last Token). The reasons may lie in the following: (1) Advantage in Holistic Metrics: In unstructured editing tasks, using the Last Token is more advantageous. Employing the Last Token in Stage 1 can incorporate richer contextual information, thereby better facilitating semantic integration in Stage 2; (2) Advantage in Fine-grained Metrics: When multiple fine-grained facts share the same Last Subject Word Token, it may trigger representational competition or conflict. The design using the Last Token helps mitigate this issue. This experiment is designed to clearly reveal the impact of different representation choices on the final editing effectiveness, thereby strengthening the persuasiveness of our design decision.

\begin{table*}[t]
\centering
\caption{The representation comparison results of Llama3-8B-Instruct on the {\benchmark-UnKE} dataset.}
\Large
\renewcommand{\arraystretch}{1} 
\resizebox{0.7\textwidth}{!}{%
\begin{tabular}{c|cc|cccc}
\toprule[1.5pt]

\multirow{2}{*}{\textbf{Mehtod}} & \multicolumn{2}{c|}{\textbf{Holistic}} & \multicolumn{4}{c}{\textbf{Fine-grained}} \\ 
\cmidrule(lr){2-7}

& \textbf{Bert-Score$\uparrow$} & \multicolumn{1}{c|}{\textbf{Rouge-L$\uparrow$}} &\textbf{Bert-Score$\uparrow$} & \textbf{Rouge-L$\uparrow$} & \textbf{HR$\uparrow$} & \textbf{\(C_{\text{LCS}}\)$\uparrow$}\\ 
\midrule[1.0pt]
\textbf{{\method} (Last Token)} & {99.36\std{0.04}} & {97.78\std{0.10}} & {65.63\std{0.16}} & {53.79\std{0.19}} & {53.31\std{0.23}} & {61.78\std{0.19}}\\
\textbf{{\method} (Last Subject Word Token)} & {98.49\std{0.06}} & {94.82\std{0.16}} & {58.44\std{0.16}} & {43.33\std{0.20}} & {42.43\std{0.23}} & {51.67\std{0.20}}\\
\bottomrule[1.5pt]

\end{tabular}
}
\label{tab:representation}
\end{table*}

\subsection{Case Study}
\label{appendix:case_study}
We present generation examples of the Llama3-8B-Instruct and Qwen2.5-7B-Instruct models after being edited by different methods, as shown in Table~\ref{case_study_llama} and Table~\ref{case_study_qwen}. Here, <Holistic> represents the model's generation result when the editing prompt is used as input, while <Fine-grained> represents the result for a fine-grained question. The text highlighted in red corresponds to the ground truth answer for that fine-grained question (which is contained within the Unstructured Target Output). Through comparative analysis, the following conclusions can be drawn:

\textbf{(1) Structured editing methods show limited performance in both holistic and fine-grained generation.} Methods represented by ROME and MEMIT fail to effectively store the unstructured target output into the model parameters, leading to factual hallucinations during generation.

\textbf{(2) Unstructured editing methods achieve good holistic generation but suffer from insufficient fine-grained knowledge recall.} Methods represented by UnKE and AnyEdit, while capable of storing and reproducing the holistic target output reasonably well, perform poorly in recalling the internal fine-grained knowledge.

\textbf{(3) {\method} demonstrates excellent performance in both holistic and fine-grained generation.} Our method not only accurately stores the unstructured target output but also effectively recalls the fine-grained knowledge contained within it, achieving stable and reliable performance in both types of generation tasks.

\begin{table*}
\footnotesize
\caption{Model Editing Case Study on Llama3-8B-Instruct}
\label{case_study_llama}
\begin{tcolorbox}[boxrule=0.5pt,left=0pt,right=0pt,top=2.5pt,bottom=2.5pt,title={Model Editing Case Study on LLAMA3}]
    \centering
    \renewcommand{\arraystretch}{2}
    \begin{tabular}{p{0.3\linewidth} p{0.6\linewidth}}
        \makecell[c]{Editing Prompt} 
         & \makecell*[{{p{8cm}}}]{What are Kathy Saltzman's notable achievements in the field of composition and what makes her a leading figure in contemporary classical music?} \\
        \hline
        \makecell[c]{Unstructured Target Output} 
         & \makecell*[{{p{8cm}}}]{Kathy Saltzman is a highly accomplished composer \ldots \ldots some of the most prestigious musical groups in the world, including \tar{the New York Philharmonic and the London Symphony Orchestra}. \ldots \ldots in contemporary classical music.} \\
        \hline
        \makecell[c]{Fine-grained Question} 
         & \makecell*[{{p{8cm}}}]{Which renowned musical groups have performed Kathy Saltzman's work?} \\
        \hline
        \hline
        \multicolumn{2}{c}{\textbf{\large Generation Output}} \\ 
        \hline
        \makecell[c]{FT-L} & \makecell*[{{p{8cm}}}]{
        \textbf{<Holistic>}: and Saltzman is a composer and composer-in-residence \ldots \ldots companies and orchestras, including
        \\
        \textbf{<Fine-grained>}: and Saltzman is a composer and composer, and the work of the composer \ldots \ldots} \\
        \hline
        \makecell[c]{ROME} & \makecell*[{{p{8cm}}}]{
        \textbf{<Holistic>}: Kathy Saltzman is not a well-known composer, and a \ldots \ldots assist you with that as well.
        \\
        \textbf{<Fine-grained>}: I apologize, but I couldn't find any information on a renowned musical group performing \ldots \ldots} \\
        \hline
        \makecell[c]{MEMIT} & \makecell*[{{p{8cm}}}]{
        \textbf{<Holistic>}: Kathy Saltzman is a renowned composer and has written \ldots \ldots numerous awards and accolades.
        \\
        \textbf{<Fine-grained>}: Kathy Saltzman is a composer and songwriter who has worked with a variety of musical \ldots \ldots} \\
        \hline
        \makecell[c]{UnKE} & \makecell*[{{p{8cm}}}]{
        \textbf{<Holistic>}: Kathy Saltzman is a highly accomplished composer \ldots \ldots some of the most prestigious musical groups in the world, including the New York Philharmonic and the London Symphony Orchestra. \ldots \ldots in contemporary classical music.
        \\
        \textbf{<Fine-grained>}: To answer your question, Kathy Saltzman is a talented musician \ldots \ldots}\\
        \hline
        \makecell[c]{AnyEdit} & \makecell*[{{p{8cm}}}]{
        \textbf{<Holistic>}: Kathy Saltzman is a highly accomplished composer \ldots \ldots some of the most prestigious musical groups in the world, including the New York Philharmonic and the London Symphony Orchestra. \ldots \ldots dedicated advocate for the arts, and
        \\
        \textbf{<Fine-grained>}: Kathy Saltzman is a talented musician, and her work has been performed \ldots \ldots}\\
        \hline
        \makecell[c]{\method} & \makecell*[{{p{8cm}}}]{
        \textbf{<Holistic>}: Kathy Saltzman is a highly accomplished composer \ldots \ldots some of the most prestigious musical groups in the world, including the New York Philharmonic and the London Symphony Orchestra. \ldots \ldots in contemporary classical music.
        \\
        \textbf{<Fine-grained>}: Some of the most prestigious musical groups in the world, including the \tar{New York Philharmonic and the London Symphony Orchestra}.}\\
    \end{tabular}
\end{tcolorbox}
\end{table*}

\begin{table*}
\footnotesize
\caption{Model Editing Case Study on Qwen2.5-7B-Instruct}
\label{case_study_qwen}
\begin{tcolorbox}[boxrule=0.5pt,left=0pt,right=0pt,top=2.5pt,bottom=2.5pt,title={Model Editing Case Study on LLAMA3}]
    \centering
    \renewcommand{\arraystretch}{2}
    \begin{tabular}{p{0.3\linewidth} p{0.6\linewidth}}
        \makecell[c]{Editing Prompt} 
         & \makecell*[{{p{8cm}}}]{What are Kathy Saltzman's notable achievements in the field of composition and what makes her a leading figure in contemporary classical music?} \\
        \hline
        \makecell[c]{Unstructured Target Output} 
         & \makecell*[{{p{8cm}}}]{Kathy Saltzman is a highly accomplished composer \ldots \ldots some of the most prestigious musical groups in the world, including \tar{the New York Philharmonic and the London Symphony Orchestra}. \ldots \ldots in contemporary classical music.} \\
        \hline
        \makecell[c]{Fine-grained Question} 
         & \makecell*[{{p{8cm}}}]{Which renowned musical groups have performed Kathy Saltzman's work?} \\
        \hline
        \hline
        \multicolumn{2}{c}{\textbf{\large Generation Output}} \\ 
        \hline
        \makecell[c]{FT-L} & \makecell*[{{p{8cm}}}]{
        \textbf{<Holistic>}: zman. It is highly likely that there is a name or a musical \ldots \ldots the work of the New Philhar
        \\
        \textbf{<Fine-grained>}: zman and her work. After a more in-depth and widely recognized music and \ldots \ldots} \\
        \hline
        \makecell[c]{ROME} & \makecell*[{{p{8cm}}}]{
        \textbf{<Holistic>}: Kathy Saltzman is a highly accomplished composer, having \ldots \ldots talent and dedication as a composer.
        \\
        \textbf{<Fine-grained>}: I do not have specific information about which musical groups have performed Kathy Saltzman's work. \ldots \ldots} \\
        \hline
        \makecell[c]{MEMIT} & \makecell*[{{p{8cm}}}]{
        \textbf{<Holistic>}: Kathy is a highly accomplished composer, having written \ldots \ldots influence composers around the world.
        \\
        \textbf{<Fine-grained>}: \ldots \ldots including \tar{the New York Philharmonic}, \tar{the London Symphony Orchestra} \ldots \ldots} \\
        \hline
        \makecell[c]{UnKE} & \makecell*[{{p{8cm}}}]{
        \textbf{<Holistic>}: Kathy Saltzman is a highly accomplished composer \ldots \ldots some of the most prestigious musical groups in the world, including the New York Philharmonic and the London Symphony Orchestra. \ldots \ldots in contemporary classical music.
        \\
        \textbf{<Fine-grained>}: \ldots \ldots including the Los Angeles Philharmonic and the Boston Pops Orchestra \ldots \ldots}\\
        \hline
        \makecell[c]{AnyEdit} & \makecell*[{{p{8cm}}}]{
        \textbf{<Holistic>}: Kathy Saltzman is a highly accomplished composer \ldots \ldots some of the most prestigious musical groups in the world, including the New York Philharmonic and the London Symphony Orchestra. \ldots \ldots dedicated advocate for the arts, and
        \\
        \textbf{<Fine-grained>}: Kathy Saltzman is a highly talented composer, having written numerous pieces \ldots \ldots}\\
        \hline
        \makecell[c]{\method} & \makecell*[{{p{8cm}}}]{
        \textbf{<Holistic>}: Kathy Saltzman is a highly accomplished composer \ldots \ldots some of the most prestigious musical groups in the world, including the New York Philharmonic and the London Symphony Orchestra. \ldots \ldots in contemporary classical music.
        \\
        \textbf{<Fine-grained>}: The \tar{New York Philharmonic} and the Los Angeles Philharmonic.}\\
    \end{tabular}
\end{tcolorbox}
\end{table*}

\end{document}